\documentclass[sigplan,screen,sigconf]{acmart}

\usepackage{tabularx}
\usepackage{array}
\newcolumntype{L}[1]{>{\raggedright\arraybackslash}p{#1}} 
\newcolumntype{Y}{>{\raggedright\arraybackslash}X}      

\setcopyright{rightsretained}
\acmDOI{}
\acmISBN{}
\acmConference[Conference '26]{}{March, 2026}{}

\begin{document}


\title{From Pragmas to Partners: A Symbiotic Evolution of Agentic High-Level Synthesis}

\author{Niansong Zhang}
\affiliation{
  \institution{Cornell University}
  \country{USA}
}

\author{Sunwoo Kim}
\affiliation{
  \institution{Cornell University}
  \country{USA}
}

\author{Shreesha Srinath}
\affiliation{
  \institution{Cerebras Systems}
  \country{USA}
}

\author{Zhiru Zhang}
\affiliation{
  \institution{Cornell University}
  \country{USA}
}

\settopmatter{authorsperrow=4}

\renewcommand{\shortauthors}{Zhang et al.}

\begin{abstract}
The rise of large language models has sparked interest in AI-driven hardware design, raising the question: does high-level synthesis (HLS) still matter in the agentic era? We argue that HLS remains essential. 
While we expect mature agentic hardware systems to leverage both HLS and RTL, this paper focuses on HLS and its role in enabling agentic optimization.
HLS offers faster iteration cycles, portability, and design permutability that make it a natural layer for agentic optimization.
This position paper makes three contributions. First, we explain why HLS serves as a practical abstraction layer and a golden reference for agentic hardware design. Second, we identify key limitations of current HLS tools, namely inadequate performance feedback, rigid interfaces, and limited debuggability that agents are uniquely positioned to address. Third, we propose a taxonomy for the symbiotic evolution of agentic HLS, clarifying how responsibility shifts from human designers to AI agents as systems advance from copilots to autonomous design partners.
\end{abstract}

\maketitle

\begin{table*}[t]
\centering
\caption{Autonomy-inspired levels for agentic HLS.}
\vspace{-.1in}
\small
\setlength{\tabcolsep}{6pt}
\renewcommand{\arraystretch}{1.15}

\begin{tabularx}{\textwidth}{c L{2.6cm} c c Y}
\hline
\textbf{Level} & \textbf{Name} & \textbf{Human Effort} & \textbf{AI Effort} & \textbf{Key Technology} \\
\hline

L0 & Manual DSE
& High & Low
& Standard HLS compilation and reports; manual code and pragma iteration; feedback and debugging largely human-driven. \\

L1 & HLS Copilot
& High & Low
& Explains HLS reports with source attribution, suggests local code and pragma edits, helps triage failures, and improves RTL readability for follow-up debugging. \\

L2 & Autotuning Agent
& Medium & Medium
& Runs closed-loop DSE over pre-defined search space using tool feedback, integrates HLS IPs using standard interfaces, generates lightweight replayable traces for performance analysis, and performs basic differential testing. \\

L3 & Guardrailed Agents
& Medium & High
& Performs multi-step tool use with strong guardrails, leverages verification to provide strong correctness guarantee, including equivalence checking, assertion synthesis, and counterexample explanations; constructs mixed-fidelity performance model, performs localized failure and bottleneck analysis for DSE. \\

L4 & Domain Architect
& Low & High
& Handles system-level optimization and integration with customized interfaces, use complete workloads to drive performance and correctness tests.\\

L5 & Silicon Partner
& Low & High
& End-to-end autonomy across architectures and tools; continual learning across projects; improves the HLS stack itself (compiler passes, scheduling heuristics, codegen, tracing and verification infrastructure). \\

\hline
\end{tabularx}
\label{tab:agentic-hls-levels}
\end{table*}

\section{Introduction}
\label{sec:intro}

High-level synthesis (HLS) lifts accelerator design from RTL to high-level software specification, but achieving high performance still requires substantial manual, iterative, expertise-driven optimization: restructuring code, rewriting kernels, and adding pragmas guided by toolchain feedback to shape compute, memory, data layout, and communication.

The rise of large language models and agentic AI has sparked renewed interest in automating this process. A natural question arises: does HLS still matter, or will AI simply generate RTL directly? We argue that HLS remains not only relevant but essential in the agentic era.
We do not claim that HLS alone or RTL alone solves agentic hardware design. We expect a mixed workflow where agents move between high-level intent, HLS source, and RTL, depending on the task. This paper focuses on the role of HLS in that workflow.

This position paper makes three contributions: (1) it argues that HLS is a strong abstraction layer for agentic hardware design, enabling iteration speed, portability, and design permutability beyond pure RTL generation (\S\ref{sec:why-hls}); (2) it highlights limitations in current HLS workflows, especially performance feedback, interfaces, verification, and debugging, that agents can help address (\S\ref{sec:limitations}); and (3) it introduces an autonomy-inspired taxonomy for the symbiotic evolution of agentic HLS, clarifying capabilities and shifting human-agent responsibilities (\S\ref{sec:taxonomy})~\cite{SAE_J3016_2021}.


\section{Why HLS Remains Relevant}
\label{sec:why-hls}

Despite advances in AI-driven RTL generation, HLS offers fundamental advantages that make it the natural abstraction layer for agentic hardware design.

\paragraph{Rapid Design Space Exploration.}
Agentic workflows require many iterations to test correctness and reach design goals/constraints. HLS enables rapid design space exploration (DSE) through two mechanisms. First, it provides an executable high-level software specification that serves as a golden reference, allowing for fast software simulation before hardware generation. Second, HLS representations are highly permutable: one design point can be easily transformed into another while preserving functional semantics. 
In contrast, direct RTL generation lacks a clear golden reference and equivalent RTL transformations require invasive rewrites that are harder to automate and more error-prone.
Operating primarily at RTL also raises the cost of iteration. RTL simulation is slower, and code and traces are more verbose, which increases the token cost for agentic loops that depend on repeated inspections and refinements. Higher abstraction is therefore even more important when optimization is driven by many agentic iterations.

\paragraph{Portability.} 
The same high-level software specification can serve as a single source to target different FPGA families, vendors, or accelerator platforms with minimal modification. Beyond platform portability, HLS enables constraint portability: the same source can be synthesized for high throughput, low latency, or low power by adjusting directives and synthesis settings. For agentic HLS systems, learned optimizations and design patterns can be transferred across platforms and constraints, enabling reuse across projects.

\paragraph{Verification and Debugging.}
HLS improves verification by centering the workflow around a readable and executable specification. The high-level source serves as a golden reference for functional validation and supports fast regression testing before synthesis. This reduces the burden of debugging in RTL while still allowing RTL checks when needed.

\section{HLS Limitations That Agents Can Address}
\label{sec:limitations}


Despite rapid DSE, portability, and ease of verification being central to the value proposition of HLS, current tools fall short in all three. We identify key gaps that agents are uniquely positioned to address.

\paragraph{Performance Feedback Limits DSE}
Fast and meaningful performance feedback is crucial for effective DSE, yet current HLS tools fall short. Synthesis reports provide latency estimates, resource utilization, and scheduling information, but these are often difficult to interpret and relate back to source code. Worse, many designs exhibit data-dependent control flow, causing tools to report unknown or variable cycle counts that provide little actionable guidance. Agents can address this through mixed-fidelity performance modeling: extracting operation schedule, loop-level information, and control-flow regions from HLS artifacts to construct performance estimates and replayable traces. This enables actionable feedback even for designs where traditional HLS reports provide only question-mark estimates, presenting insights at the same abstraction level where designers make optimization decisions.

\paragraph{Rigid Interface Limits Portability.}
A major pain point is composing accelerators with the surrounding system. Tools such as Vitis HLS~\cite{VitisHLS} support standard interfaces such as AXI, but offer limited ability to customize or extend these protocols. Integrating HLS-generated IP with systems built using different frameworks, such as the LiteX SoC builder~\cite{Kermarrec2020LiteX}, requires manual writing interface adapters and bridge logic. 
Tools such as Catapult HLS~\cite{CatapultHLS} support more flexible interfaces, but defining protocols remains a complex task.
This \textit{composability gap} undermines the portability promise of HLS. AI agents can bridge this gap by understanding both the HLS design intent and the target system's interface requirements, then generating the necessary adapter logic and protocol bridges.

\paragraph{Lack of End-to-End Verification and Debugging.}
HLS verification today is still split between high-level software simulation,
RTL simulation, and ad hoc debugging, with limited help connecting failures back to source intent. Agents can use HLS as a golden reference for both humans and RTL, enabling differential testing, equivalence checking, assertion synthesis, and targeted counterexample explanation across abstraction levels. For many workloads, we also want testbenches written at an even higher level, such as PyTorch models, to drive software simulation and RTL co-simulation with realistic inputs and oracle outputs. This integration is not available in current flows, but an agent could mediate it by generating adapters, harnesses, and simulation drivers.
Finally, agents can improve the readability of the generated RTL by refactoring the structure, naming signals, and adding documentation, which reduces the cost of debugging when designers must resort to RTL.

\section{Levels of Automation}
\label{sec:taxonomy}

Recent work explores LLM assistance for HLS, including code generation and guidance~\cite{Xiong2024HLSPilot,Liao2024AreLLMsGoodHLS}, automated repair~\cite{Xu2024HLSRepair}, retrieval-augmented optimization~\cite{Xu2024RALAD}, software-to-hardware bridging~\cite{DBLP:journals/todaes/ColliniGK25}, and agentic DSE and multi-agent design~\cite{Sheikholeslam2024SynthAI,Oztas2024AgenticHLS,Wang2025LLMDSE}. These systems span a wide range of autonomy and human involvement. To characterize this spectrum of autonomy and human involvement, we propose the taxonomy in Table~\ref{tab:agentic-hls-levels}. The taxonomy is useful not just for labeling systems, but for highlighting what must improve in agentic HLS to enable greater autonomy, especially actionable performance feedback, flexible interface, improved verification, and debuggability.

At \textbf{L0 (Manual DSE)}, designers iterate manually from HLS reports. \textbf{L1 (HLS Copilot)} assists with report interpretation and localized edit suggestions, but humans still implement and validate. \textbf{L2 (Autotuning Agent)} runs closed-loop optimization over a predefined space and automates routine evaluation and integration. \textbf{L3 (Guardrailed Agents)} performs multi-step tool use with strong guardrails, with verification and mixed-fidelity feedback for targeted debugging and DSE. \textbf{L4 (Domain Architect)} shifts humans to intent definition and review, while agents drive system-level optimization and workload-level validation within a bounded domain. Finally, \textbf{L5 (Silicon Partner)} extends autonomy across architectures and tools, including improving the HLS stack itself.

Current systems operate primarily at L1--L2, with research pushing toward L3. Advancing further depends directly on addressing the composability and feedback limitations discussed in \S\ref{sec:limitations}.

\section{Conclusion}
\label{sec:conclusion}


We argue that HLS will remain a valuable layer in the agentic era, supporting faster iteration, portability, and design exploration. We highlight performance feedback, rigid interfaces, and debuggability as key gaps in today’s HLS workflows where agents can help, and propose a taxonomy of automation levels to structure the space. We hope that this paper encourages joint progress between the HLS and AI communities.

\newpage

\bibliographystyle{ACM-Reference-Format}
\bibliography{ref}

\end{document}